\documentclass[lettersize,journal]{IEEEtran}
\usepackage{amsmath,amsfonts}
\usepackage{algorithmic}
\usepackage{algorithm}
\usepackage{array}
\usepackage[caption=false,font=normalsize,labelfont=sf,textfont=sf]{subfig}
\usepackage{textcomp}
\usepackage{stfloats}
\usepackage{url}
\usepackage{booktabs}
\usepackage{verbatim}
\usepackage{graphicx}
\usepackage{xcolor}
\usepackage{color}
\usepackage{cite}
\usepackage{amsmath}
\usepackage{multirow}
\usepackage{pifont}
\usepackage{bbding}
\usepackage{fontawesome}

\newcommand{\xmark}{\ding{55}}%

\newcommand{\z}{\mathbf{z}}
\begin{document}

\title{Variational Inference for Bird's Eye View Segmentation in Autonomous Driving}

\author{
Jingyue Shi, Huaicheng Li, Junhui Zhao,~\IEEEmembership{Senior Member,~IEEE}, Yanxiang Jiang

\thanks{Supported by the Fundamental Research Funds for the Central Universities 2025JBZX060, National Engineering Research Center of System Technology for High-Speed Railway and Urban Rail Transit under Grant 2024YJ255 and National Natural Science Foundation of China (U25B2007). (Corresponding author: Junhui Zhao.) 

Jingyue Shi, Huaicheng Li and Junhui Zhao are with School of Electronic and Information Engineering, Beijing Jiaotong University, Beijing 100044, China (e-mail: 22115004@bjtu.edu.cn, 23115040@bjtu.edu.cn, junhuizhao@hotmail.com).

Yanxiang Jiang is with School of Information Science and Engineering, Southeast University, Nanjing 210096, China (e-mail: yxjiang@seu.edu.cn).

}
}

\markboth{IEEE Transactions on Circuits and Systems for Video Technology}%
{Shell \MakeLowercase{\textit{et al.}}: A Sample Article Using IEEEtran.cls for IEEE Journals}


\maketitle

\begin{abstract}
The bird’s eye view (BEV) has emerged as a pivotal approach for environmental perception in autonomous driving, providing a unified spatial representation for vehicles. Nevertheless, despite BEV’s significance in addressing the challenges inherent to autonomous driving, effectively fusing data from multiple camera sensors and operating in complex external driving environments remains a considerable challenge. To mitigate this issue, we recast the BEV segmentation problem within a variational inference framework. In this paper, we propose a novel transformer-based variational flow transformation network for BEV segmentation, denoted as TVB. Our architecture implicitly learns the mapping from multiple camera views to a unified canonical BEV map during training by exploiting posterior BEV supervision. TVB employs a conditional variational auto-encoder (CVAE) as its backbone and produces multiple BEV map candidates. To augment the realism of the generated BEV maps, we integrate normalizing flows into the map generation process, enabling the construction of more complex and expressive probability distributions. Furthermore, we design a BEV-attention fusion (BAF) module that harnesses attention mechanisms to adaptively integrate the multiple candidate BEV maps. Experimental results, evaluated on both the nuScenes and OPV2V datasets, demonstrate that our proposed method achieves superior performance in multi-camera view BEV segmentation and lane environment perception.
\end{abstract}

\begin{IEEEkeywords}
Autonomous driving, BEV maps, condition variational auto-encoder, BEV segmentation, normalizing flow.
\end{IEEEkeywords}

\section{Introduction}
\begin{figure}[t]  
\begin{center}
   \includegraphics[width=1\linewidth]{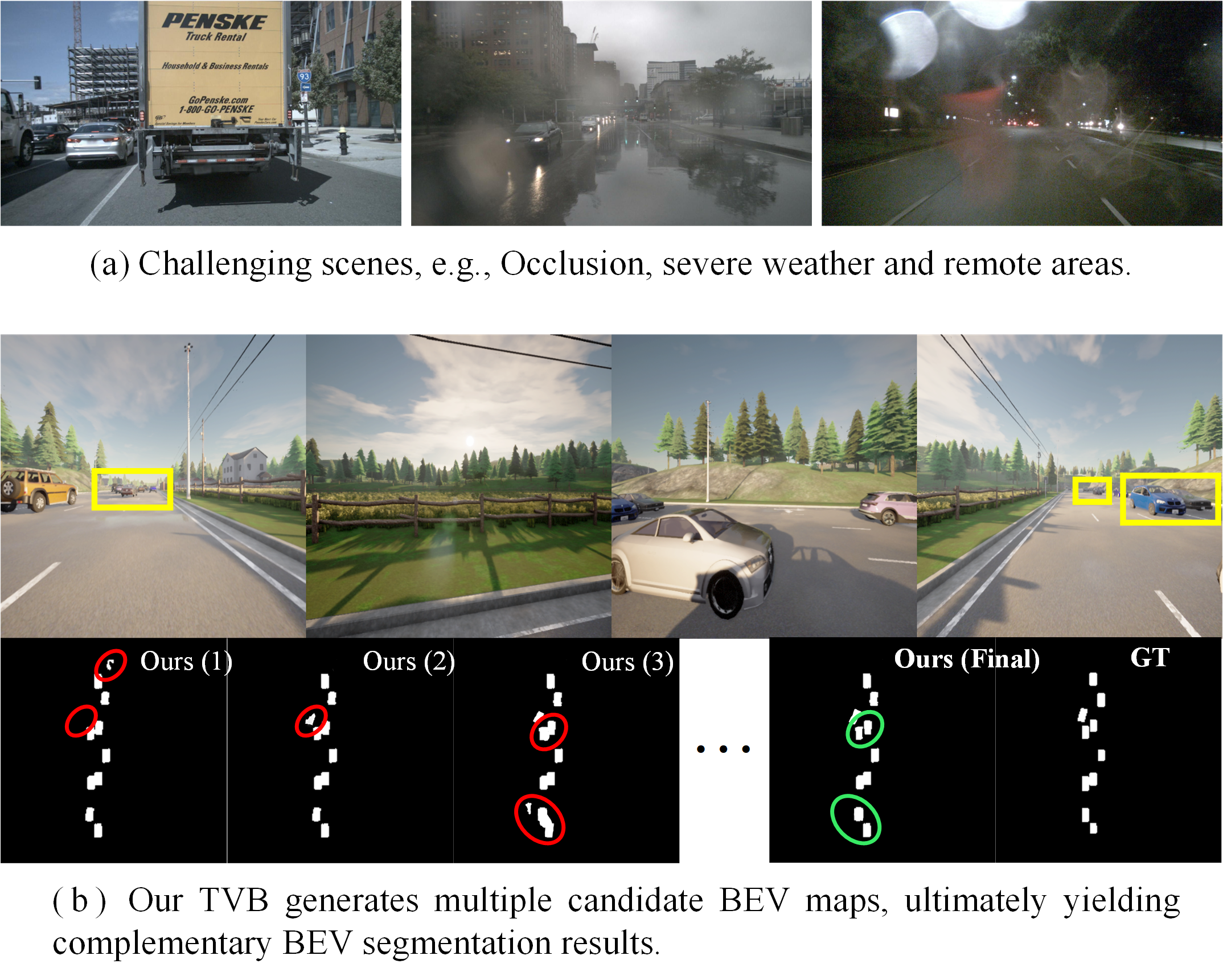}  
\end{center}
\caption{The motivation and intention of our proposed method. (a) Some challenging scenes. (b) In the first row are the multiple camera views in the vehicle, with the yellow boxes representing remote areas and occlusion. In the second row, Ours (1), Ours (2), and Ours (3) are the three candidate predictions for TVB, where the red ellipse is the wrong BEV result. Ours (Final) is the weighted average of all candidate predictions.}
\label{Fig1}
\end{figure}

\IEEEPARstart{B}{ird's} eye view (BEV) refers to a perspective typically obtained from directly above or from a high-altitude vantage point, which is useful for observing scenes, maps, or environments \cite{xu2026tigdstill, yang2025ralibev}. The central concept involves fusing spatial features from one or multiple views and projecting them onto a self-referenced BEV space. BEV \cite{zhao2024bev} is particularly significant in robotics and autonomous driving \cite{yang2025drivemoe} because it provides a comprehensive global perspective, enabling systems to understand their surroundings, facilitate effective path planning \cite{jia2023towards}, and enhance both safety and efficiency. In the context of Vehicle-to-Vehicle (V2V) communication~\cite{zhao2019computation,zhao2020edge}, vehicles can collaboratively develop more advanced autonomous driving solutions by exchanging and sharing BEV information. However, a major challenge remains in effectively acquiring, integrating, and interpreting large volumes of sensor data~\cite{9693129} from complex traffic scenarios, such as occlusion \cite{wang2024toward} and remote areas \cite{9786052} (as illustrated in Fig.~\ref{Fig1}(a)). These scenarios adversely affect the extraction of features from detected objects, potentially resulting in the generation of erroneous BEV representations, which, in turn, compromises the reliable operation of autonomous driving systems in real-world settings.

BEV semantic segmentation is considered a crucial method for obtaining BEV \cite{zhou2026UniSparseBEV}. Traditional methods \cite{mallot1991inverse,reiher2020sim2real} typically employ the inverse perspective mapping (IPM) function to compute the internal and external parameters of the camera, mapping features from the perspective space to the BEV space. This process involves camera calibration and perspective transformation. After mapping, segmentation and analysis are performed in the BEV. While IPM is a useful tool for bridging BEV spatial views, it can cause visual distortions when mapping images or features to the ground plane. This can hinder accurate object localization in complex scenarios with occlusions or long-range views \cite{qiao2024local}. Traditional IPM methods may exhibit suboptimal performance in such cases.

Recently, CNN-based approaches have gradually emerged as the mainstream methods for BEV semantic segmentation. These approaches can be broadly categorized into explicit and implicit methods. Explicit methods \cite{wang2019pseudo, ma2019accurate} typically incorporate depth information to determine the field of view. In particular, certain techniques convert pixels from the view into point clouds by utilizing depth data, subsequently processing these point clouds for downstream tasks. However, these methods are constrained by the inherent sparsity of point clouds and challenges associated with network generalization. To address these limitations, some approaches explicitly predict the distribution of depth information and guide the dispersion of 2D features to corresponding 3D positions through the outer product of image feature maps and depth distributions. Although these methods are relatively efficient, their ability to generate accurate BEV features in dynamic environments can be compromised in obscured or distant regions, leading to imprecise results.

Implicit methods \cite{lu2019monocular, pan2020cross} frequently employ architectures such as the multi-layer perceptron (MLP) \cite{10128757} and transformer \cite{vaswani2017attention} networks to generate BEV representations through a data-driven approach. MLP-based techniques primarily map features from single monocular images onto the BEV space for subsequent fusion, yet they often inadequately incorporate geometric information in regions of overlapping views, leading to a loss of local feature correlations. Transformer-based methods \cite{zhou2022cross, bartoccioni2023lara} reformulate the BEV generation problem as a sequence-to-sequence transformation. By establishing a learnable reverse query in the BEV, these approaches employ attention mechanisms to search for highly correlated image features across multiple views, thereby achieving notable results. However, for distant targets, occlusion, and perception in complex environments, information queries are needed for a higher degree of grid refinement in the BEV features. This undoubtedly increases the computational load and training difficulty of the network.

To address the challenges associated with obtaining accurate BEV representations in complex scenarios and to overcome the limitations of existing models, we propose an alternative generative model: the transformer-based variational flow transformation BEV segmentation network (TVB). TVB first integrates information from multiple camera sensors to extract BEV features and then employs a conditional variational auto-encoder (CVAE) \cite{sohn2015learning, zhao2017learning} as the backbone network to generate multiple BEV maps. These maps are sampled in the variational latent space and are intelligently fused into the final BEV representation through adaptive merging mechanisms facilitated by attention, as shown in Fig.~\ref{Fig1}(b). In addition, to ensure that the generated BEV maps more closely resemble the true instances, we incorporate normalizing flow transformations \cite{papamakarios2021normalizing} into the CVAE (CFLOW). During the training phase, cross-attention mechanisms are first used to fuse views from multiple camera sensors. The fused BEV features, along with their corresponding ground truth BEV, are fed into the posterior network, while, in parallel, the prior network receives only the BEV features as input. Both networks encode these inputs and map them into their respective latent spaces in order to capture the structural distribution of the BEV. The Kullback-Leibler (KL) divergence is used to approximate the transition from the prior distribution to the posterior distribution. However, using an axis-aligned Gaussian latent distribution in all CVAE-based models may not accurately approximate the posterior distribution when it exhibits non-linear structures, multimodal distributions, or complex boundaries \cite{kingma2016improved}. To obtain a more complex and expressive latent distribution, we introduce normalizing flow into the posterior network. We then perform multiple random samples in the transformed latent distribution and feed them into the decoder to reconstruct BEV maps, which are subsequently fused to obtain the ultimate BEV.During the inference phase, the network takes only images from multiple camera sensors as input. The encoder transforms these images into the latent space and performs multiple samplings. The sampled codes are then fed into the decoder to reconstruct BEV maps, which are finally fused to produce the final BEV representation.

To identify the optimal BEV from multiple BEV maps, a common approach is to sum them and then compute their average.  Although effective for producing a final BEV, this method may inadvertently overlook individual BEV maps that are potentially more accurate. To address this problem, we propose a BEV-attention fusion (BAF) module that adaptively integrates multiple BEV maps. Thus far, our TVB framework has demonstrated an enhanced ability to exploit the CFLOW backbone, resulting in the generation of multiple BEV maps and yielding more robust and accurate BEV outcomes.

Overall, our main contributions are as follows:

\begin{itemize}
    \item In this paper, we propose a novel transformer-based variational BEV segmentation network, denoted as TVB, which incorporates CFLOW (CVAE with Normalizing Flow) to model expressive latent BEV distributions.
    \item We incorporate normalizing flow into CVAE to generate multiple more expressive candidate BEV maps. In addition, we develop a BEV-attention fusion (BAF) module to adaptively merge multiple complementary BEV maps to produce accurate and robust BEV results.
    \item We evaluate the TVB on the OPV2V \cite{xu2022opv2v} and nuScenes \cite{caesar2020nuscenes} datasets, and the experiments show that our TVB achieves state-of-the-art performance in BEV segmentation. We also validate the importance of modules in the network during ablation experiments.

\end{itemize}

\section{Related works}

\subsection{BEV Semantic Segmentation} 
BEV segmentation is critical for supplying essential perceptual information needed for subsequent behavior planning and prediction in autonomous driving and robotics. In most BEV segmentation approaches, features are typically extracted from single or multiple views and subsequently transformed into BEV. Early work often relied on IPM‐based BEV segmentation \cite{senguptaautomatic}, which addresses mathematically intractable problems by imposing additional constraints. However, the points in the inverse mapping are positioned on a horizontal plane, and IPM heavily depends on the assumption of a flat ground.To mitigate these issues, some later methods attempted to improve the view transformation process. For example, TrafCam3D \cite{zhu2021monocular} introduced a robust homography mapping network for dual-view images to enhance the detection accuracy of distorted BEV images. OGMs \cite{loukkal2021driving} transformed the segmentation results of objects projected from a perspective view onto the ground plane into BEV, thereby avoiding distortions caused by the vehicle body. Cam2BEV \cite{reiher2020sim2real} projected features from multiple in-vehicle camera images onto the BEV to yield an integrated BEV semantic map. Although these methods alleviate some limitations of IPM, they still fall short of meeting the demands of autonomous driving in complex real-world scenarios.
\begin{figure*}[t]  
\begin{center}
   \includegraphics[width=1\linewidth]{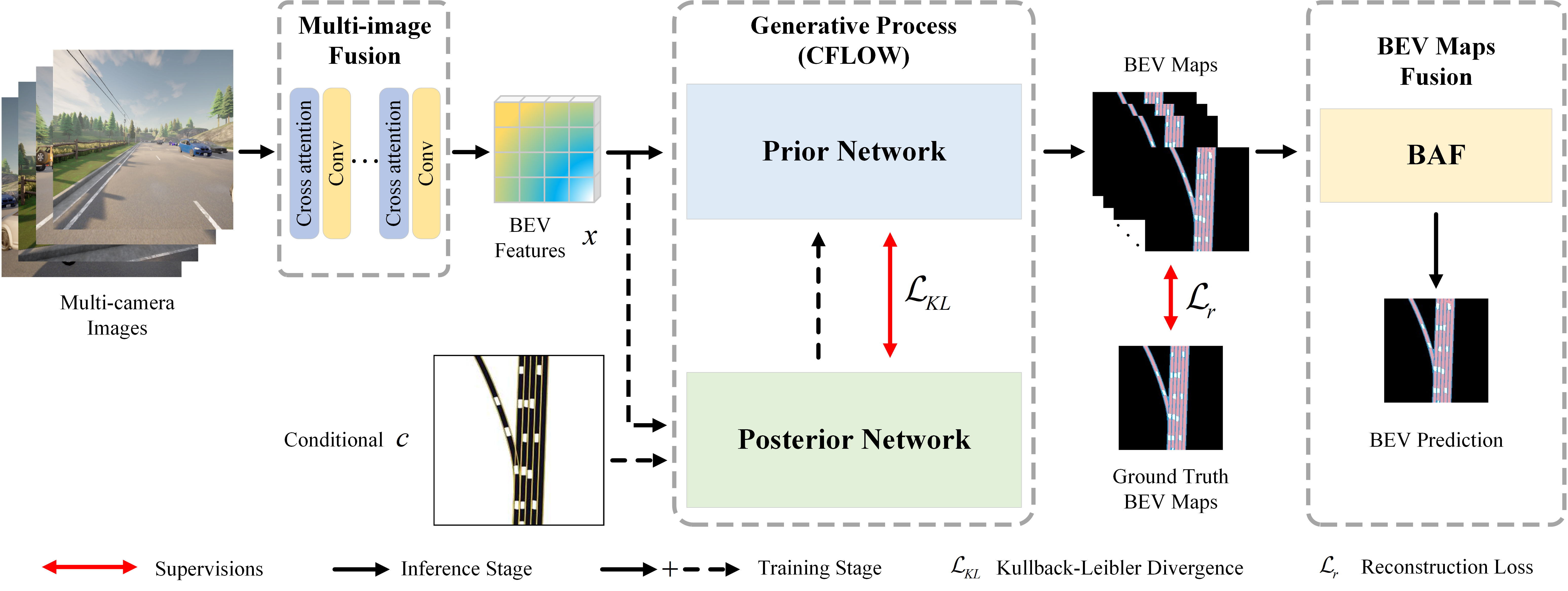}  
\end{center}
\caption{The architecture of TVB. During the training phase, TVB first fuses multiple camera images from different viewpoints to obtain BEV features \textbf{x}. It then feeds \textbf{x} and the ground truth values \textbf{c} of BEV into CFLOW. KL divergence is utilized to minimize the distance between the prior latent distribution and the posterior latent distribution, ultimately yielding multiple candidate BEV maps. In the inference phase, multiple camera images from different viewpoints serve as the sole input to the network, producing multiple BEV maps as output. The maps obtained in both phases undergo adaptive fusion through the BAF module to obtain the final BEV segmentation.}
\label{Fig2}
\end{figure*}
Another line of methods makes full use of depth information, elevating 2D pixels or features into 3D space. Pseudo-LiDAR \cite{wang2019pseudo} proposed a method to combine the original image with depth information, transforming it into a pseudo-LiDAR that replaces the original LiDAR point cloud to represent the BEV. Lift-Splat \cite{philion2020lift} predicts the category distribution of the depth and context feature vectors. The outer product defines the features of points on perspective rays, and it fuses the features of multiple views to make the network more robust to camera calibration errors. BEVDet \cite{huang2021bevdet} improved upon Lift-Splat \cite{philion2020lift} and introduced a 3D object detection network under multiple camera views, including an image encoder, a view transformer, a BEV encoder, and a detection head. However, due to the fact that most of these methods perceive from a top-down perspective, they are unable to accurately capture vertical information, such as the height and depth of objects, which may lead to occlusion problems and perception difficulties in complex environments.

Inspired by Transformer \cite{vaswani2017attention}, attention mechanisms have been increasingly applied to multi-view fusion tasks, yielding advanced results. For instance, CVT \cite{zhou2022cross} embeds both the intrinsic and extrinsic calibration of each camera into map queries to fuse images from multiple sensors. FlatFusion \cite{zhu2025flatfusion} proposes a camera-Lidar fusion scheme based on sparse Transformer. It delves deeply into the details of fusion by sparsifying and flattening the features of the two modalities in a unified 3D space. BEVSegFormer \cite{peng2023bevsegformer} employs a common backbone to encode features from images captured by any camera, subsequently enhancing these features using a deformable transformer encoder. Additionally, BEVFormer \cite{li2022bevformer} replaces conventional multi-head attention with deformable attention to more effectively integrate features from multiple views with dense queries, while also leveraging temporal cues to incorporate historical BEV information into BEV queries. Similarly, CoBEVT \cite{xu2022cobevt} introduces an innovative attention mechanism, Fusion Axis Attention (FAX), which effectively extracts region-specific details and contextual cues at a reduced computational cost. BEVDiffuser \cite{ye2025bevdiffuser} adds noise to the feature maps generated by the existing BEV model and uses the real target layout as a conditional guide to recover a clean BEV representation from the noisy feature maps through the diffusion model learning. In contrast to approaches that introduce additional overhead, our study exclusively relies on cross-attention mechanisms to merge images from different camera sensors for BEV feature extraction. These BEV features subsequently serve as conditional inputs for the CVAE backbone network.

\subsection{Variational Auto-Encoder} 
Variational auto-encoders (VAE) \cite{kingma2013auto,rezende2014stochastic} rank among the most widely adopted generative models, fundamentally relying on the transformation of input data into latent variables. This process encompasses learning the distribution of the latent space, sampling from it, and ultimately decoding the samples to generate new data. However, because VAEs are trained in an unsupervised manner, generating samples specifically tailored to particular tasks remains challenging. In response, the conditional variational auto-encoder (CVAE) \cite{sohn2015learning} has been introduced for representation learning and structured prediction, demonstrating excellent performance in segmentation and semantic labeling by conditionally generating images with specific attributes. For example, \cite{yan2016attribute2image} employed CVAE to learn latent variables representing the foreground and background in facial images, thereby generating faces with a designated skin tone. Similarly, \cite{esser2018variational} developed a U-Net-based CVAE capable of producing images with diverse appearances under specific body poses. Moreover, \cite{huang2023gaze} utilized CVAE to address the limited availability of eye features in complex environments by generating graphical representations of eye images to enhance the accuracy of 3D gaze direction predictions. BEV-VAE \cite{chen2025bev} builds a unified and spatially aligned BEV latent space based on CVAE, achieving high-quality, consistent and controllable generation of multi-view images in autonomous driving scenarios. To acquire the BEV representation under specific conditions, our work leverages BEV features as conditional information for the CVAE backbone network, generating multiple candidate BEV maps.

\subsection{Normalizing Flow}
Normalizing flows employ a sequence of computationally efficient Jacobi \cite{papamakarios2021normalizing,kingma2016improved} bijective transformations (i.e., a flow chain) to map simple base distributions onto more complex ones. Owing to their reversible nature, flow-based generative models can accurately capture the distribution of real data, including latent variables, thereby enabling precise log-likelihood estimation. For instance, SRFlow \cite{lugmayr2020srflow} introduces a super-resolution method based on regularized flows, which leverages learned image posterior probabilities for image processing. Similarly, PointFlow \cite{yang2019pointflow} uses continuous normalizing flows to model the distribution of points by mapping samples drawn from a simple prior distribution (e.g., a 3D Gaussian) into points that constitute the target shape, ultimately producing high-fidelity 3D point clouds. Although the CVAE represents a latent variable model, it only approximates the actual distribution and cannot precisely quantify the discrepancy between the latent distribution and the true one, a limitation that can lead to blurred images in complex scenes. In this study, we integrate flow normalization into the CVAE framework to generate more realistic BEV maps.

\section{Method}
TVB is organized into two distinct stages, as depicted in Fig.~\ref{Fig2}. In the first stage, CFLOW learns and produces BEV representations, yielding several candidate outputs. In the subsequent stage, these candidate BEV representations are processed by the adaptive fusion module, which generates the final BEV segmentation.

During the training phase, we first fuse camera images from multiple views into BEV features \textbf{x} using cross-attention. Subsequently, \textbf{x} is fed into the encoder of the posterior network along with the ground truth \textbf{c} of the BEV to obtain latent codes containing both types of information. At the same time, \textbf{x} is fed into another prior network to obtain latent codes containing \textbf{x} information. The KL divergence is then used to assess the similarity between the two codes. Multiple sampling is then performed on the latent codes from the prior network, which are then fed into the decoder to reconstruct of BEV maps. Finally, the BAF module is used to fuse multiple candidate BEV maps to obtain the BEV segmentation.
\begin{figure}[t]  
\begin{center}
   \includegraphics[width=1\linewidth]{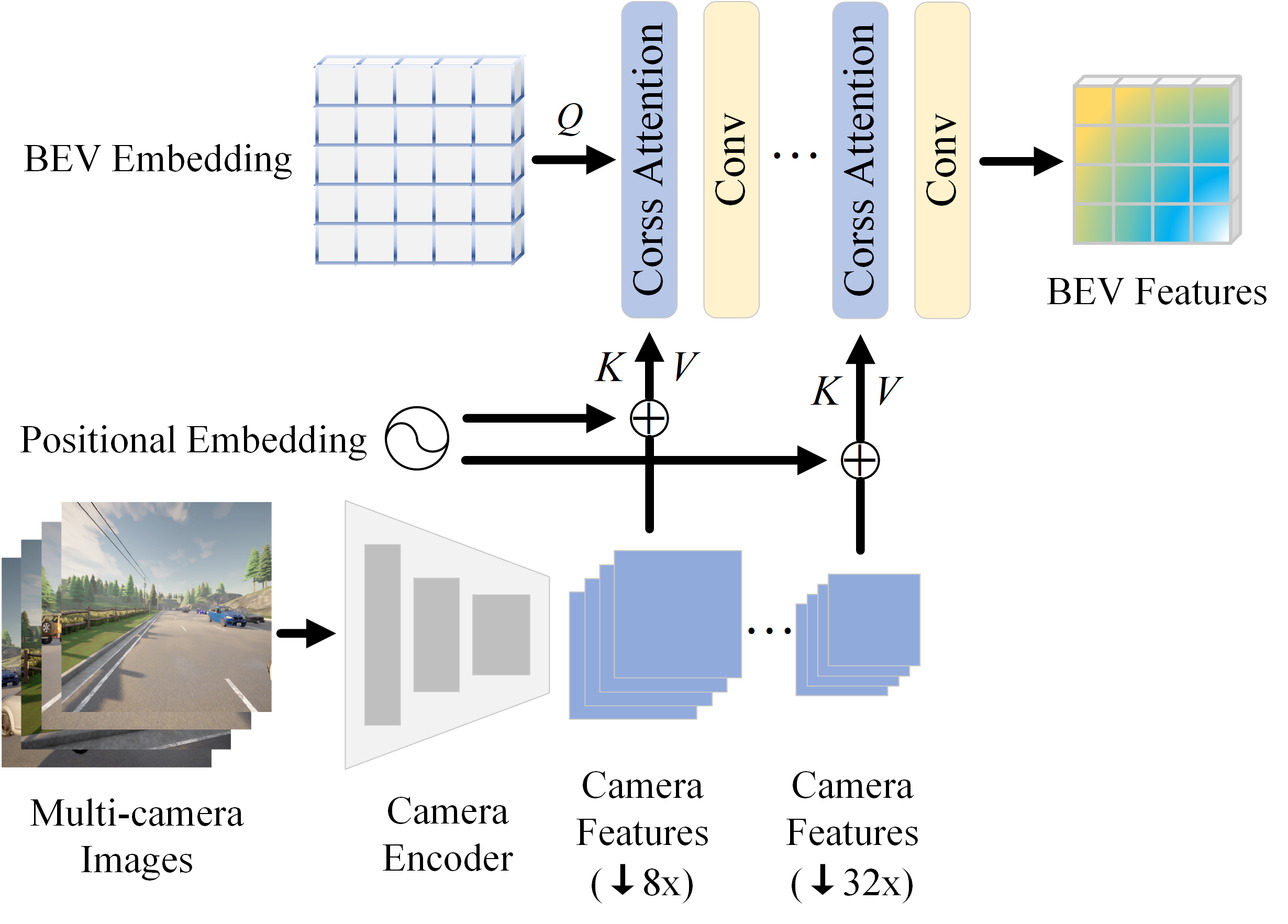}  
\end{center}
\caption{Multiple views fusion architecture. Multiple camera images from different viewpoints are fed into an encoder, which extracts features at different scales for each image. The camera's position embedding is obtained based on its intrinsic parameters, and a BEV embedding is initialized as a query. Multiple cross-attention mechanisms are employed to learn the position embeddings for each view and to refine the BEV features.}
\label{Fig3}
\end{figure}
\subsection{Multi-View Fusion}
The network takes a set of $j$ monocular views from cameras $\{{{I}_{i}}\in {{\mathbb{R}}^{H\times W\times 3}},{{K}_{i}}\in {{\mathbb{R}}^{3\times 3}},{{R}_{i}}\in {{\mathbb{R}}^{3\times 3}},{{t}_{i}}\in {{\mathbb{R}}^{3}}\}_{i=1}^{j}$ as input, where $I_{i}$, $K_{i}$, $R_{i}$, and $t_{i}$ represent the input image, camera intrinsics, extrinsic rotation, and translation, respectively. The network calculates the BEV feature $\textbf{F} \in \mathbb{R}^{H \times W \times C}$ (with height $H$, width $W$, and channels $C$) through cross-view information fusion. This \textbf{F} serves as the input for subsequent CFLOW to generate BEV representation. The processing architecture for this BEV is similar to CVT \cite{zhou2022cross}, as shown in Fig.~\ref{Fig3}.

In detail, data from multiple views are initially processed by a CNN to extract feature representations at various scales. These extracted features subsequently undergo a linear transformation to yield the value vector employed in the attention mechanism. Furthermore, the camera intrinsic and extrinsic parameter matrices from each view are incorporated into the multi-scale features for position encoding, thereby serving as the key vector in the attention mechanism for refinement. Additionally, a high-resolution BEV embedding is initialized as the query in the attention mechanism. In contrast to the low-resolution queries used in CVT, the high-resolution query enables a more focused extraction of small object features. Finally, the multi-level feature maps output by the CNN are cascaded to further refine the BEV features.

\subsection{BEV Representation}
Instead of directly segmenting BEV from its features, we utilize CFLOW to construct the BEV representation. Before delving into the application of conditional variational inference combined with normalizing flows for BEV learning, we first offer a succinct overview of the VAE. VAE, a generative model \cite{kingma2013auto, rezende2014stochastic}, is designed to generate new data samples by capturing the underlying probability distribution of the data. It comprises an encoder and a decoder, operating as follows: the encoder transforms the input data \textbf{x} into a probability distribution over the latent space, ${q}_{\phi }(\textbf{z}|\textbf{x})$, where \textbf{z} represents the latent variable. Typically, this distribution is Gaussian. The encoder's objective is to learn the mean and variance that map the input data into the latent space. A sample is drawn from the probability distribution learned by the encoder, denoted as $\textbf{z} \sim {p}_{\theta}$, which then serves as the latent representation of the input. Subsequently, the decoder converts this latent point back into the data space, generating samples similar to the original input. The VAE is optimized by maximizing the evidence lower bound (ELBO) of the log-likelihood.
\begin{align}
   \log {{p}_{\theta }}(\textbf{x})\ge -{{D}_{KL}}[{{q}_{\phi }}(\textbf{z}|\textbf{x})||{{p}_{\theta }}(\textbf{z})]+{{\mathbb{E}}_{{{q}_{\phi }}(\textbf{z}|\textbf{x})}}[\log {{p}_{\theta }}(\textbf{x}|\textbf{z})],
\label{eq1}
\end{align}
where ${{D}_{KL}}$ is the KL divergence used to measure the difference between two probability distributions. During the gradient descent optimization process, the reparameterization trick \cite{kingma2013auto} is applied to sample from the latent space.
\begin{figure*}[t!]  
\begin{center}
   \includegraphics[width=1\linewidth]{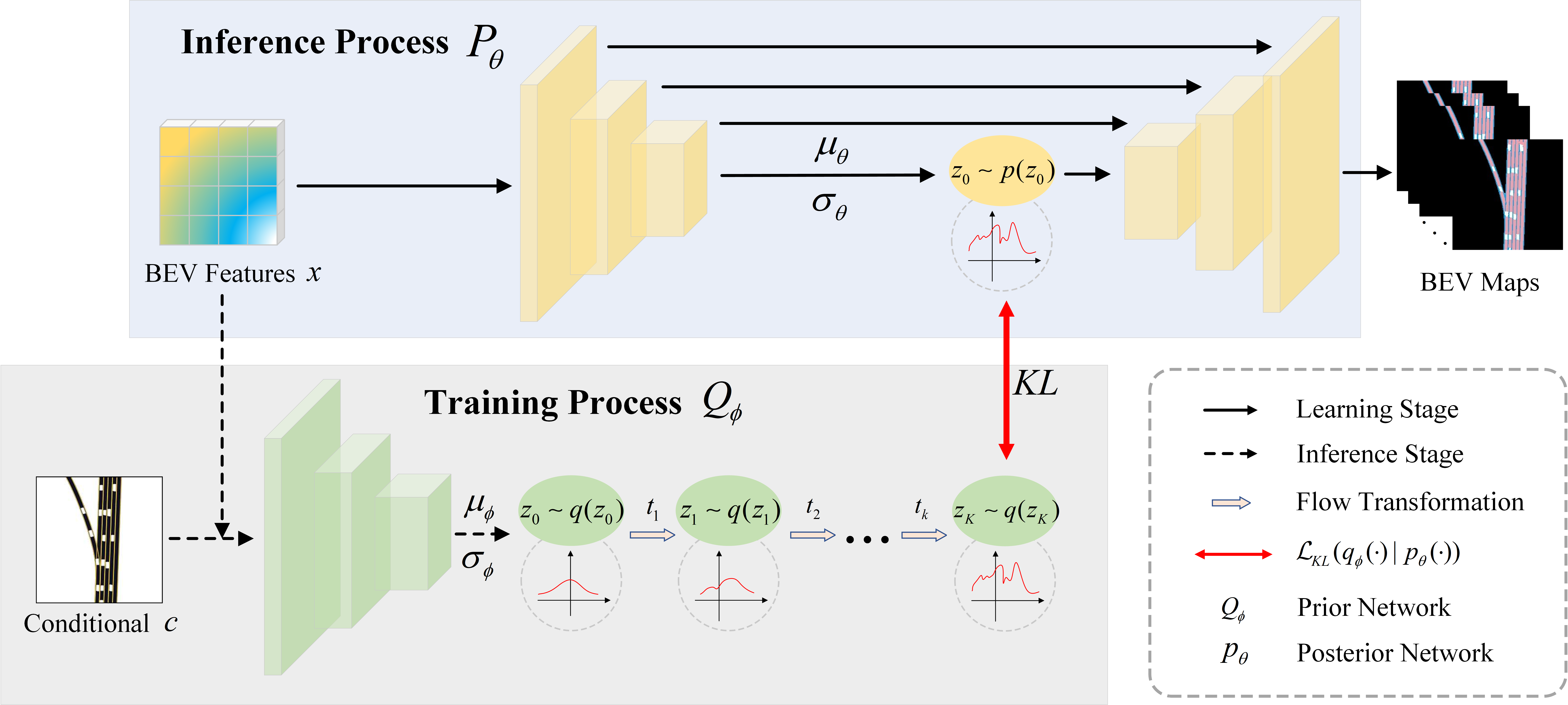}  
\end{center}
\caption{The architecture of the CFLOW network. (a) The training process consists of solid and dashed lines: the prior network and posterior network obtain the latent prior distribution ${{p}_{\theta }}({{z}_{0}}|x)$ and latent posterior distribution ${{q}_{\phi }}({{z}_{0}}|c,x)$, and the KL loss is employed to minimize the distance between these distributions. The distribution ${{p}_{\theta }}({{z}_{0}}|x)$ is transformed into a more expressive distribution ${{p}_{\theta }}({{z}_{K}}|x)$ through $K$ normalizing flow steps in CFLOW. Finally, multiple BEV maps are sampled from ${{p}_{\theta }}({{z}_{K}}|x)$. (b) The inference process, represented by solid lines, involves inputting BEV features into the prior network, undergoing the normalizing flow transformations in CFLOW}, and sampling to obtain multiple BEV maps.
\label{Fig4}
\end{figure*}

Due to the over smoothing and difficulty in distinguishing changes under different conditions in samples generated by VAE, we incorporate conditional information, the ground truth (GT) \textbf{c} of BEV, into VAE to form a conditional variational framework for supervised learning of BEV representation. For this purpose, by conditioning the \textbf{c} on the input BEV feature \textbf{x}, we can obtain the variational posterior $\log p(\textbf{c}|\textbf{x})$ as follows:
\begin{align}
   \log p(\textbf{c}|\textbf{x})& =\log {{\int }_{\textbf{z}}}p(\textbf{c},\textbf{z}|\textbf{x})d\textbf{z}\ge {{\mathbb{E}}_{q}}\left[ \log \frac{p(\textbf{c},\textbf{z}|\textbf{x})}{q(\textbf{z}|\textbf{x},\textbf{c})} \right] \notag\\ 
 & ={{\mathbb{E}}_{q}}\left[ \log \frac{p(\textbf{c}|\textbf{z},\textbf{x})p(\textbf{z}|\textbf{x})}{q(\textbf{z}|\textbf{x},\textbf{c})} \right],
\label{eq2}
\end{align}
here, we denote by $\log p(\textbf{c}|\textbf{x})$ the log-likelihood of the BEV map conditional on the multi-view fused BEV feature \textbf{x}, which needs to be maximized in the context of Bayesian inference. Meanwhile, $\log \int_{\textbf{z}} p(\textbf{c},\textbf{z}|\textbf{x})d\textbf{z}$ represents the log marginal probability of the joint distribution of \textbf{c} and the latent variable \textbf{z} given \textbf{x}, where we integrate over all possible values of \textbf{z}. Next, Jensen's inequality is applied to convert this integral of the log probability into an expectation, indicated by ${{\mathbb{E}}_{q}}[\cdot]$, where the expectation is computed with respect to \textbf{z} following the distribution $q(\textbf{z}|\textbf{x},\textbf{c})$. Furthermore, we divide the log joint probability by the log probability of the approximate distribution. The term $p(\textbf{c},\textbf{z}|\textbf{x})$ is then further decomposed into $p(\textbf{c}|\textbf{z},\textbf{x})$, the conditional distribution within the generative model, and $p(\textbf{z}|\textbf{x})$, which serves as the prior distribution.distribution in the generative model, and $p(\textbf{z}|\textbf{x})$ is the prior distribution.

Therefore, we further rewrite the ELBO for conditional inference as:
\begin{align}
   {{\mathcal{L}}_{ELBO}}={{\mathbb{E}}_{q}}[\log p(\textbf{c}|\textbf{z},\textbf{x})]-{{D}_{KL}}[q(\textbf{z}|\textbf{x},\textbf{c})||p(\textbf{z}|\textbf{x})].
\label{eq3}
\end{align}

Thus, we obtain a conditional prior, which is different from the uninformative prior in VAE.

\subsection{CFLOW: CVAE with Normalizing Flow}
For generative models with structures similar to CVAE, the generated images often lack diversity, resulting in blurred and similar images. This issue is particularly evident in the probabilistic U-Net model employed for segmentation tasks \cite{baumgartner2019phiseg, kohl2019hierarchical}. In this model, the prior distribution ${p}_{\theta}$ is usually modeled as a standard normal distribution $\mathcal{N}(0,I)$, representing a relatively basic structure. Consequently, when the posterior distribution ${q}_{\phi}$ is applied to approximate the prior, the limited flexibility of the chosen prior may hinder ${q}_{\phi}$ from effectively capturing more complex variations present in the data. To address this issue, we introduce normalizing flow into CVAE. Normalizing flow is a method of constructing more complex and expressive probability distributions through a series of computationally tractable invertible transformations (flow steps). The flow $T$ consists of $K$ steps, each of which is an invertible transformation. These transformations convert a simple distribution $p({\textbf{z}}_{0})$ into a more complex and expressive distribution $p({\textbf{z}}_{K})$, and the transformation process is given by $T={T}_{K}\circ {T}_{K-1}\circ \cdots {T}_{1}$.

This enables the model to more accurately capture the diverse range of dynamic vehicle and static road information present in complex scenarios. The transformation process for the forward flow ${{\textbf{z}}_{0}}\to {{\textbf{z}}_{K}}$ and the inverse flow ${{\textbf{z}}_{K}}\to {{\textbf{z}}_{0}}$ is expressed mathematically as:
\begin{align}
  & {{\textbf{z}}_{k}}={{T}_{k}}({{\textbf{z}}_{k-1}})\text{ for }k=1\ldots K, \notag\\ 
 & {{\textbf{z}}_{k-1}}={{T}_{k}}^{-1}({{\textbf{z}}_{k}})\text{ for }k=K\ldots 1,
\label{eq4}
\end{align}
they denote the transformations from a simple to a complex distribution and vice versa, respectively. The initial latent variable, denoted as ${{\textbf{z}}_{0}}$, follows a distribution specified by $p({{\textbf{z}}_{0}})$. For an individual flow step indexed by $k$, the probability distribution $p({{\textbf{z}}_{K}})$ can be written as the product of $p({{\textbf{z}}_{K-1}})$ and the reciprocal of the Jacobian determinant, typically expressed as:
\begin{align}
  p({{\textbf{z}}_{k}})& =p({{\textbf{z}}_{k-1}}){{\left| \frac{\partial {{T}_{k}}({{\textbf{z}}_{k-1}})}{\partial {{\textbf{z}}_{k-1}}} \right|}^{-1}} \notag\\ 
 & =p({{\textbf{z}}_{k-1}}){{\left| {{J}_{{{T}_{k}}}}({{\textbf{z}}_{k-1}}) \right|}^{-1}},
\label{eq5}
\end{align}
here, ${{J}_{{{T}_{k}}}}$ represents the Jacobian determinant. Therefore, the transformation process for the entire normalizing flow in the log domain is
\begin{align}
  \log p({{\textbf{z}}_{K}})=\log p({{\textbf{z}}_{0}})+\log \left| \prod\limits_{k=1}^{K}{J_{{{T}_{k}}}^{-1}({{\textbf{z}}_{k-1}})} \right|,
\label{eq6}
\end{align}
the latter term in the equation includes the cumulative sum of the absolute value of the logarithm of the Jacobian determinant for each flow step $k$. We can further rewrite Eq. (\ref{eq6}) as:
\begin{align}
  \log p({{\textbf{z}}_{K}})=\log p({{\textbf{z}}_{0}})-\sum\limits_{k=1}^{K}{\log |{{J}_{{{T}_{k}}}}({{\textbf{z}}_{k-1}})|}.
\label{eq7}
\end{align}
\begin{figure*}[!th]  
\begin{center}
   \includegraphics[width=0.8\linewidth]{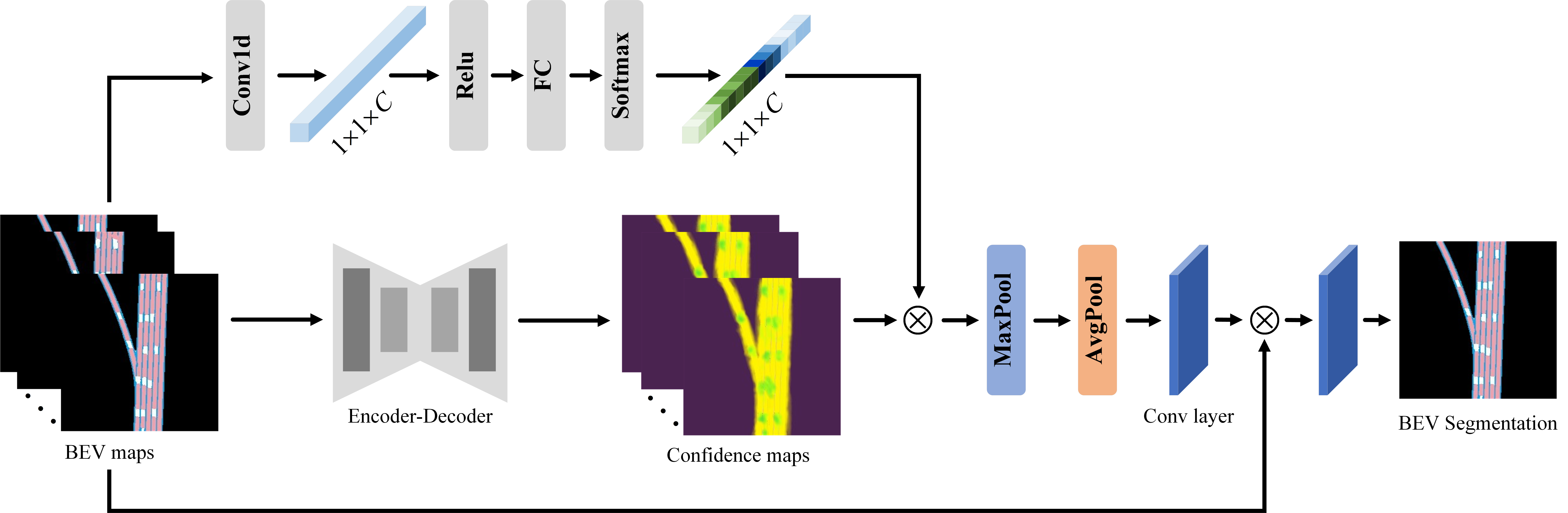}  
\end{center}
\vspace{-2mm}
\caption{The framework of the proposed BEV-attention fusion module. Multiple confidence maps are obtained from various BEV maps, and an attention mechanism is utilized to allocate different weights to them. Global information aggregation is applied to the fused confidence map to obtain the final BEV segmentation.}
\label{Fig5}
\vspace{-2mm}
\end{figure*}
In Eq. (\ref{eq3}), the optimization standard objective in CVAE is the negative of the conditional log-likelihood limit. With a normalizing flow of $K$ steps, we can use the first-order Markov assumption to decompose the $k$-step random variable as follows:
\begin{align}
  \log q({{\mathbf{z}}_{K}}|\mathbf{c},\mathbf{x})=\log \prod\limits_{k=1}^{K}{q({{\mathbf{z}}_{k}}|{{\mathbf{z}}_{k-1}},\mathbf{c},\mathbf{x})}. 
\label{add1}
\end{align}

The transformation density of each step can be obtained from the flow transformation in Eq. (\ref{eq4}):
\begin{align}
  \log q({{\mathbf{z}}_{K}}|\,\mathbf{x})=\log q({{\mathbf{z}}_{k-1}}|\mathbf{c},\mathbf{x})+\log \left| {{J}_{{{T}_{k}}}}({{\mathbf{z}}_{k-1}}|\mathbf{x}) \right|. 
\label{add2}
\end{align}

The posterior distribution $q({{\textbf{z}}_{0}}|\textbf{c},\textbf{x})$ composed of latent variables ${{\textbf{z}}_{0}}$ can be obtained from the CVAE posterior network. This distribution $q({{\mathbf{z}}_{0}}|\mathbf{c},\mathbf{x})$ can then be made more expressive using the normalizing flow method described above. Following Eq. (\ref{add2}) and Eq. (\ref{eq4}), the stream chain transformation of the underlying distribution is performed in $K$ steps, so that the transformed distribution of $\mathbf{z}$$={{\mathbf{z}}_{K}}$ can be modified from Eq. (\ref{add2}) to yield:
\begin{align}
  \log q({{\textbf{z}}_{K}}|\textbf{c},\textbf{x})=\log q({{\textbf{z}}_{0}}|\textbf{c},\textbf{x})-\sum\limits_{k=1}^{K}{\log |{{J}_{{{T}_{k}}}}({{\textbf{z}}_{k-1}}|\textbf{x})|}.
\label{eq8}
\end{align}
where we designate the transformed variable at step $K$ as the posterior variable of interest (i.e., $\z={{\z}_{K}}$). By employing this flow transform distribution from Eq. (\ref{eq8}), the optimization objective after integrating Flow into CVAE can be formulated as:
\begin{align}
  {{\mathcal{L}}_{CFLOW}}=& -{{\mathbb{E}}_{{{q}_{\phi }}({{\textbf{z}}_{0}}|\textbf{c},\textbf{x})}}[\log {{p}_{\theta }}(\textbf{c}|{{\textbf{z}}_{K}},\textbf{x})] \notag\\ 
 & \text{         }+{{D}_{KL}}[{{q}_{\phi }}({{\textbf{z}}_{0}}|\textbf{c},\textbf{x})||{{p}_{\theta }}({{\textbf{z}}_{K}}|\textbf{x})] \notag\\ 
 & \text{         }-{{\mathbb{E}}_{{{q}_{\phi }}({{\textbf{z}}_{0}}|\textbf{c},\textbf{x})}}\left[ \sum\limits_{k=1}^{K}{\log |{{J}_{{{T}_{k}}}}({{\textbf{z}}_{k-1}}|\textbf{x})|} \right],
\label{eq9}
\end{align}
here, ${{\textbf{z}}_{k}}\sim \mathcal{N}(\mu _{k}^{post}({{\textbf{z}}_{k}},\textbf{x},\textbf{c}),\sigma _{k}^{post}({{\textbf{z}}_{k}},\textbf{x},\textbf{c}))$, with $\mu _{k}^{post}\in {{\mathbb{R}}^{H\times W}}$ and $\sigma _{k}^{post}\in {{\mathbb{R}}^{H\times W}}$ denoting the mean and variance of the conditional Gaussian distribution, respectively. The first component quantifies the discrepancy between the model-predicted conditional distribution and the true distribution, and it can also be regarded as a reconstruction loss, expressed as ${{\mathcal{L}}_{r}}=\frac{1}{N}\sum\limits_{n=1}^{N}{||{{c}_{n}}-}{{\hat{c}}_{n}}||_{2}^{2}$, where $N$ is the total number of generated samples. The second term represents the KL divergence between the prior and the conditional posterior distributions, and is denoted by ${{\mathcal{L}}_{KL}}$. Finally, the third term is the expectation computed after applying the CFLOW transformation to the base distribution ${{q}_{\phi }}({{\textbf{z}}_{0}}|\textbf{c},\textbf{x})$, which we refer to as ${{\mathcal{L}}_{J}}$. Consequently, we can simplify Eq. (\ref{eq9}) as:
\begin{align}
  {{\mathcal{L}}_{CFLOW}}={{\lambda }_{1}}{{\mathcal{L}}_{r}}+{{\lambda }_{2}}{{\mathcal{L}}_{KL}}+{{\lambda }_{3}}{{\mathcal{L}}_{J}},
\label{eq10}
\end{align}
here, $\lambda$ represents the hyperparameter that weights the importance of each term.

As shown in the Fig.~\ref{Fig4}, CFLOW is divided into two components: the prior network ${{P}_{\theta}}$ and the posterior network ${{Q}_{\phi}}$. The prior network ${{P}_{\theta}}$ comprises an encoder ${{E}_{\theta}}$ and a decoder ${{D}_{\theta}}$, both of which are constructed from \textit{m} residual blocks \cite{he2016identity} featuring 3×3 convolution kernels with a stride of 1. To facilitate effective information exchange between the down-sampling and up-sampling stages, skip connections are established between ${{E}_{\theta}}$ and ${{D}_{\theta}}$. In contrast, the posterior network ${{Q}_{\phi}}$ includes only an encoder ${{E}_{\phi}}$, which has the same structure as ${{E}_{\theta}}$. In the posterior distribution, CFLOW applies a \textit{K}-step normalizing flow to further transform the latent distribution so that it better approximates the true data distribution. Subsequently, the KL divergence is used to align the prior distribution generated by the encoder ${{E}_{\theta}}$with the posterior distribution. Finally, the samples extracted from the transform prior distribution are reconstructed into the BEV map through the decoder ${{D}_{\theta}}$. It is noteworthy that the posterior network ${{Q}_{\phi}}$ is utilized exclusively during the training phase, while only the prior network ${{P}_{\theta}}$ is active during inference.

\begin{figure}[t!]  
\begin{center}
   \includegraphics[width=1\linewidth]{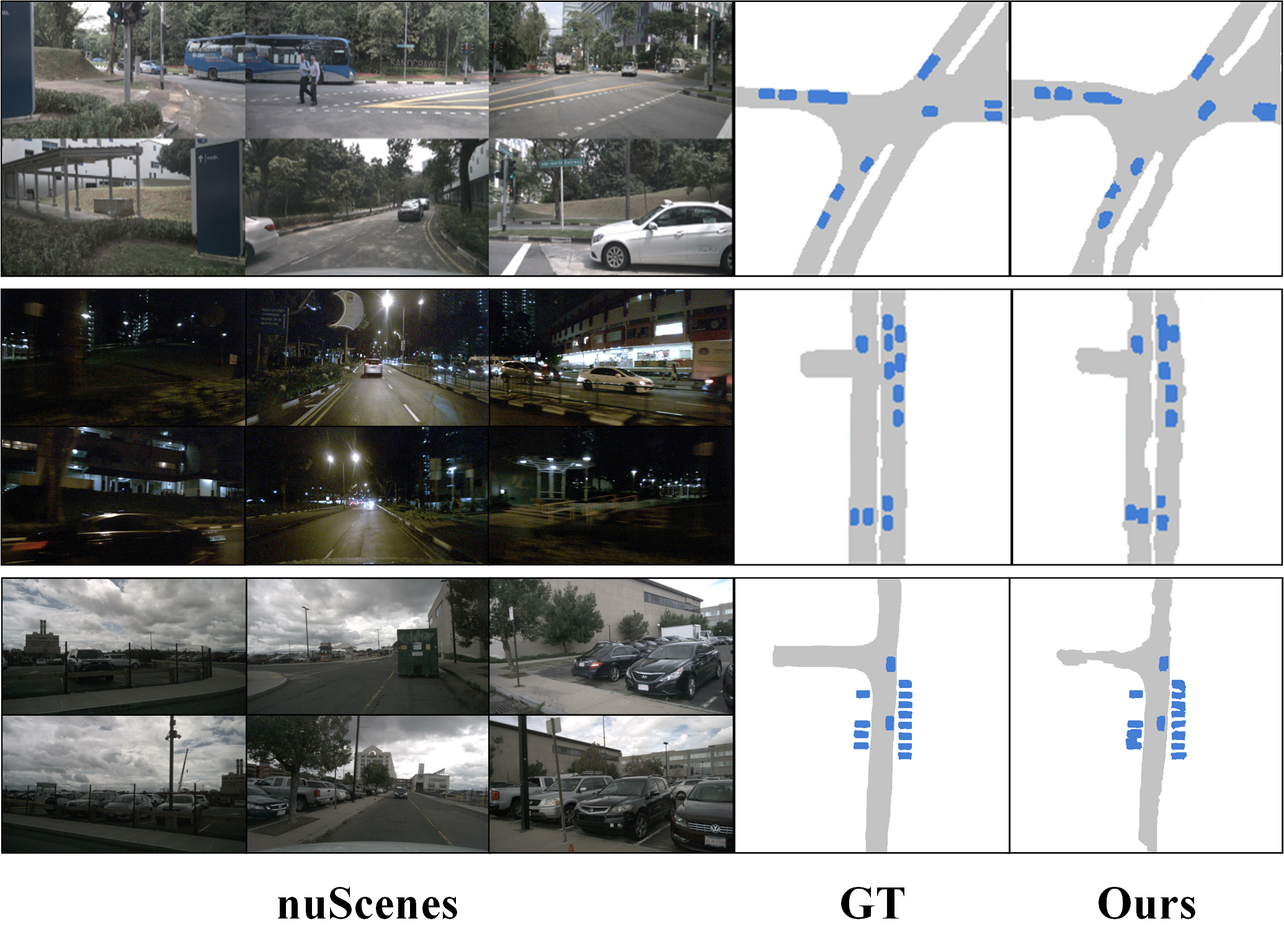}  
\end{center}
\caption{Visualization examples of BEV segmentation results of TVB on the nuScenes validation set.}
\label{Fig6}
\end{figure}
\vspace{-4mm}
\subsection{BEV-attentive Fusion}
To integrate multiple BEV maps, we propose a BAF module, as shown in Fig.~\ref{Fig5}. During the BEV segmentation prediction process, we generate multiple BEV maps ${{\hat{c}}_{n}}\in {{\mathbb{R}}^{h\times w}}$ using Monte Carlo (MC) sampling, where $h$ represents the height and $w$ represents the width. To achieve the best prediction results, we first predict the confidence map ${{\hat{f}}_{n}}$ for the BEV maps. We then use an attention mechanism to determine the contribution of each BEV map to its corresponding confidence map.
\begin{align}
  G_{n}^{S}=\text{Softmax(W}_{M}^{{{S}_{2}}}\text{RELU(W}_{M}^{{{S}_{1}}}{{\hat{c}}_{n}}\text{)),}
\label{eq11}
\end{align}
here, $G_{n}^{S}\in {{\mathbb{R}}^{1\times 1}}$, $m=h\times w$, $\text{W}_{M}^{{{S}_{1}}}\in {{\mathbb{R}}^{m\times 1}}$ and $\text{W}_{M}^{{{S}_{2}}}\in {{\mathbb{R}}^{m\times m}}$ are two learnable matrices. Subsequently, the obtained weights $G_{n}^{S}$ are used to fuse with the confidence map ${{\hat{f}}_{n}}$, i.e., $\hat{f}=\frac{1}{N}\sum\nolimits_{n=1}^{N}{{{{\hat{f}}}_{n}}G_{n}^{S}}$.

\begin{figure}[t!]  
\begin{center}
   \includegraphics[width=1\linewidth]{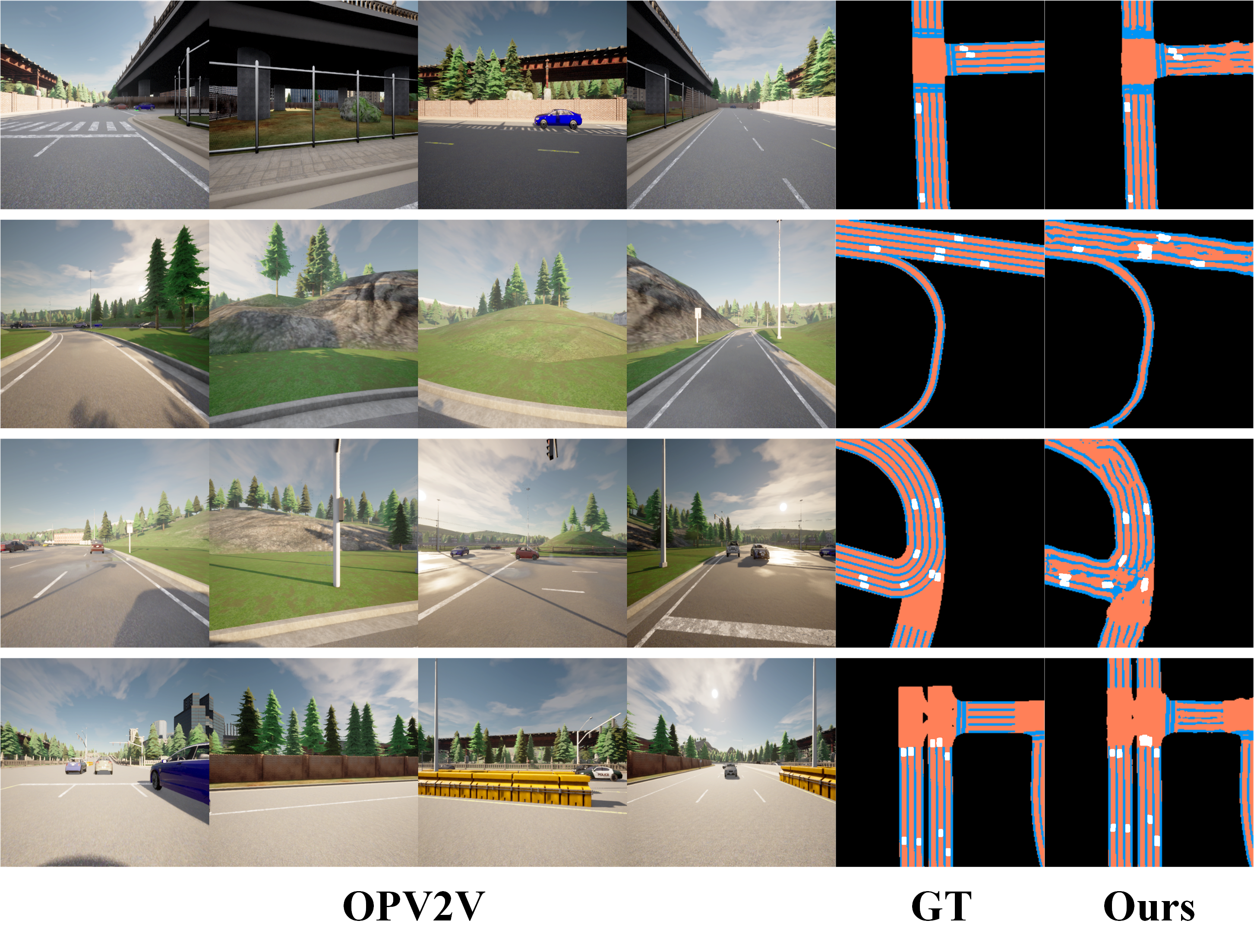}  
\end{center}
\caption{Visualization examples of BEV segmentation results of TVB on the OPV2V validation set.}
\label{Fig7}
\end{figure}
We aim to emphasize the importance of different locations by focusing on different spatial positions of the fused confidence map. To achieve this, we employ a spatial attention mechanism similar to those used in previous studies. This allows the model to concentrate its attention on specific regions of the input data, resulting in a more localized. This approach allows the model to flexibly handle spatial structures in images or sequences and balance information from different locations. Specifically, we perform successive average pooling and max pooling operations on $\hat{f}$ to highlight effective feature descriptions. Subsequently, we generate a spatial attention map through convolutional layers
\begin{align}
  L(\hat{f})=\text{RELU(Conv(}[\text{}Avg\text{(}\hat{f}\text{);}Max\text{(}\hat{f}\text{) }]\text{)),}
\label{eq12}
\end{align}
here, $L(\hat{f})\in {{\mathbb{R}}^{1\times h\times w}}$, Conv represents the convolution operation, $Avg$ and $Max$ represent average pooling and max pooling operations, respectively. We embed the obtained spatial attention map into the BEV maps to get
\begin{align}
  \hat{\textbf{y}}=\text{Conv(\textit{L}(}\hat{f}\text{)}\cdot {{\hat{c}}_{n}}\text{),}
\label{eq13}
\end{align}
here, $\hat{\textbf{y}}\in {{\mathbb{R}}^{C\times H\times W}}$ represents the final BEV segmentation prediction. Finally, we apply the Intersection-over-Union (IoU) metric to measure the degree of overlap between our predicted segmentation results and the corresponding ground truth values
\begin{align}
  {{\mathcal{L}}_{\text{IoU}}}\text{(}\hat{\textbf{y}},{{\textbf{y}}_{g}}\text{)=}\left| \frac{(\hat{\textbf{y}}\bigcap {{\textbf{y}}_{g}})}{(\hat{\textbf{y}}\bigcup {{\textbf{y}}_{g}})} \right|,
\label{eq14}
\end{align}
where the ${{\textbf{y}}_{g}}\in {{\mathbb{R}}^{C\times H\times W}}$ is represents the ground truth values.

\section{Experiments}
\subsection{Datasets}
We evaluated the performance of the proposed TVB on two benchmarks, OPV2V \cite{xu2022opv2v} and nuScenes \cite{caesar2020nuscenes}.

The nuScenes dataset comprises 40,000 frames sampled across 1,000 scenes, providing a large-scale autonomous driving dataset. Each scene, lasting roughly 20 seconds, contains 40 frames. Six cameras installed on the vehicle capture a complete 360° view, and the evaluation targets the ego vehicle's surroundings, spanning an area of $\text{100m}\times \text{100m}$ with a map resolution set at 0.5m.

The OPV2V dataset is a large dataset designed for collaborative perception research. It includes data from two simulators, CARLA and OpenCDA. CARLA provides perception data for vehicles, while OpenCDA simulates realistic traffic flow. The dataset comprises 73 scenes, demonstrating traffic flow approaching realism within 25 seconds. The dataset includes scenes from nine different cities, covering six road types. Eight of the cities were provided by CARLA, while one city was reconstructed from Culver City, Los Angeles, with real traffic flow. Each frame in the dataset contains 3D point clouds and RGB data from multiple self-driving cars. The data is captured by four cameras that cover a 360° view. Our experiments only used the camera data, and we evaluated the surrounding area of the ego vehicle with a map resolution of 0.39m.
\begin{table*}[ht]
\renewcommand\arraystretch{1.6}
\caption{Comparison of results of different methods on the nuScenes dataset. Our settings follow FIERY \cite{hu2021fiery}, “Setting 1” means 100m × 50m at 25cm resolution, and “Setting 2” means 100m × 100m at 50cm resolution. Both settings are measured by the Intersection-over-Union (IoU), with higher IoUs representing better performance. We also give the FPS results of TVB compared with other methods. Here, the number of parameters of our method is when the flow transformation step $K$ = 4.}
\begin{center}
\setlength{\tabcolsep}{3.3mm}{
\begin{tabular}{c|ccccc}
\toprule
\multirow{2}{*}{Method} & \multicolumn{5}{c}{IoU}                                                                              \\ 
                        & \multicolumn{1}{c}{Setting 1} & \multicolumn{1}{c}{Setting 2 (Vehicle)}  & \multicolumn{1}{c}{Setting 2 (Drivable Area)}& \multicolumn{1}{c}{\#Params (M)} & FPS \\ \hline
VPN \cite{pan2020cross} & \multicolumn{1}{c}{25.5}          & \multicolumn{1}{c}{-}      & \multicolumn{1}{c}{-} & \multicolumn{1}{c}{18}         &-     \\ 
STA \cite{saha2021enabling}& \multicolumn{1}{c}{36.0}          & \multicolumn{1}{c}{-}      & \multicolumn{1}{c}{-} & \multicolumn{1}{c}{-}         &-     \\ 
Lift-Splat \cite{philion2020lift}& \multicolumn{1}{c}{-}          & \multicolumn{1}{c}{32.1}   & \multicolumn{1}{c}{72.9} & \multicolumn{1}{c}{14}         &25     \\ 
FIERY \cite{hu2021fiery} & \multicolumn{1}{c}{37.7}          & \multicolumn{1}{c}{35.8}        & \multicolumn{1}{c}{-} & \multicolumn{1}{c}{7}         &8     \\ 
CVT \cite{zhou2022cross} & \multicolumn{1}{c}{37.5}          & \multicolumn{1}{c}{36.0}         & \multicolumn{1}{c}{74.3} & \multicolumn{1}{c}{\textbf{5}}         &\textbf{35}     \\ 
LaRa \cite{bartoccioni2023lara}& \multicolumn{1}{c}{41.7}          & \multicolumn{1}{c}{39.6}       & \multicolumn{1}{c}{75.4} & \multicolumn{1}{c}{7.6}         &19     \\  
Simpb \cite{tang2024simpb}& \multicolumn{1}{c}{41.4}          & \multicolumn{1}{c}{39.9}       & \multicolumn{1}{c}{75.2} & \multicolumn{1}{c}{-}         &-     \\ \hline
Ours                    & \multicolumn{1}{c}{\textbf{43.2}}          & \multicolumn{1}{c}{\textbf{41.3}}  & \multicolumn{1}{c}{\textbf{76.6}}& \multicolumn{1}{c}{7.8}         &19    \\ \bottomrule
\end{tabular}}
\end{center}
\label{tab1}
\end{table*}
\subsection{Train Details}
We resized the input images to 512×512, and during the process of fusing images from different camera views, we used ResNet34 to obtain three feature maps at different scales—(64,64), (32,32), and (16,16), corresponding to downsampling rates of $8\text{x}$, $16\text{x}$, and $32\text{x}$, respectively. The cross-attention mechanism used an initial BEV query size of 128×128×128 for the OPV2V dataset and 100×100×32 for the nuScenes dataset. In CFLOW, the encoder and decoder of the prior network consist of $m=4$ residual blocks, and the structure of the encoder of the posterior network is consistent with the prior network. The normalizing flow transformation involves a sequence of $K=4$ steps, with the latent space dimension set to 6. Throughout the model training process, the Adam optimizer was used together with a cosine annealing learning rate scheduler initialized at 0.0002. The model underwent training for 90 epochs on the OPV2V dataset and 150 epochs on the nuScenes dataset. The implementation was done using the PyTorch framework, and all experiments were executed on an RTX 4090 GPU with a total batch size of 16.

\subsection{Comparison With Prior Work}
We evaluated TVB on the nuScenes and OPV2V datasets and compared it with previous methods. On the nuScenes dataset, we selected VPN \cite{pan2020cross}, STA \cite{saha2021enabling}, Lift-Splat \cite{philion2020lift}, FIERY \cite{hu2021fiery}, CVT \cite{zhou2022cross} and LaRa \cite{bartoccioni2023lara} for comparison, TVB achieved the best IoU results in “Setting 1” (+1.7), “Setting 2 (Vehicles)” (+1.4) and “Setting 2 (Drivable Area)” (+1.4), as shown in Table \ref{tab1}. We visualized the BEV segmentation results of TVB on the nuScenes dataset in Fig.~\ref{Fig6}. On the OPV2V dataset, we compared TVB with CVT \cite{zhou2022cross}, F-Cooper \cite{chen2019f}, AttFuse \cite{xu2022opv2v}, V2VNet \cite{wang2020v2vnet}, and DiscoNet \cite{li2021learning}, achieving the best performance in three categories (vehicles: +3.3, driveable area: +0.6, lane: +2.3), and the results are presented in Table \ref{tab2}. In addition, we show the visualization of BEV segmentation in Fig.~\ref{Fig7}. The experimental results demonstrate that TVB exhibits excellent performance, providing accurate BEV segmentation results even in challenging scenarios. In Fig.~\ref{Fig9}, we show the results of TVB visualizations in night and rainy scenarios. Due to occlusion and remote view, certain features of vehicles and driveable areas in the camera image are obscured. Consequently, CVT \cite{zhou2022cross} and LaRa \cite{bartoccioni2023lara} produce inaccurate BEV segmentation. In contrast, our TVB generates multiple candidates and adaptively fuses them to obtain “Ours (Final)”. The specifics of TVB's reconstruction of BEV are depicted in the green box under occlusion and remote view, demonstrating the effectiveness and robustness of TVB.
\begin{figure*}[t!]  
\begin{center}
   \includegraphics[width=0.95\linewidth]{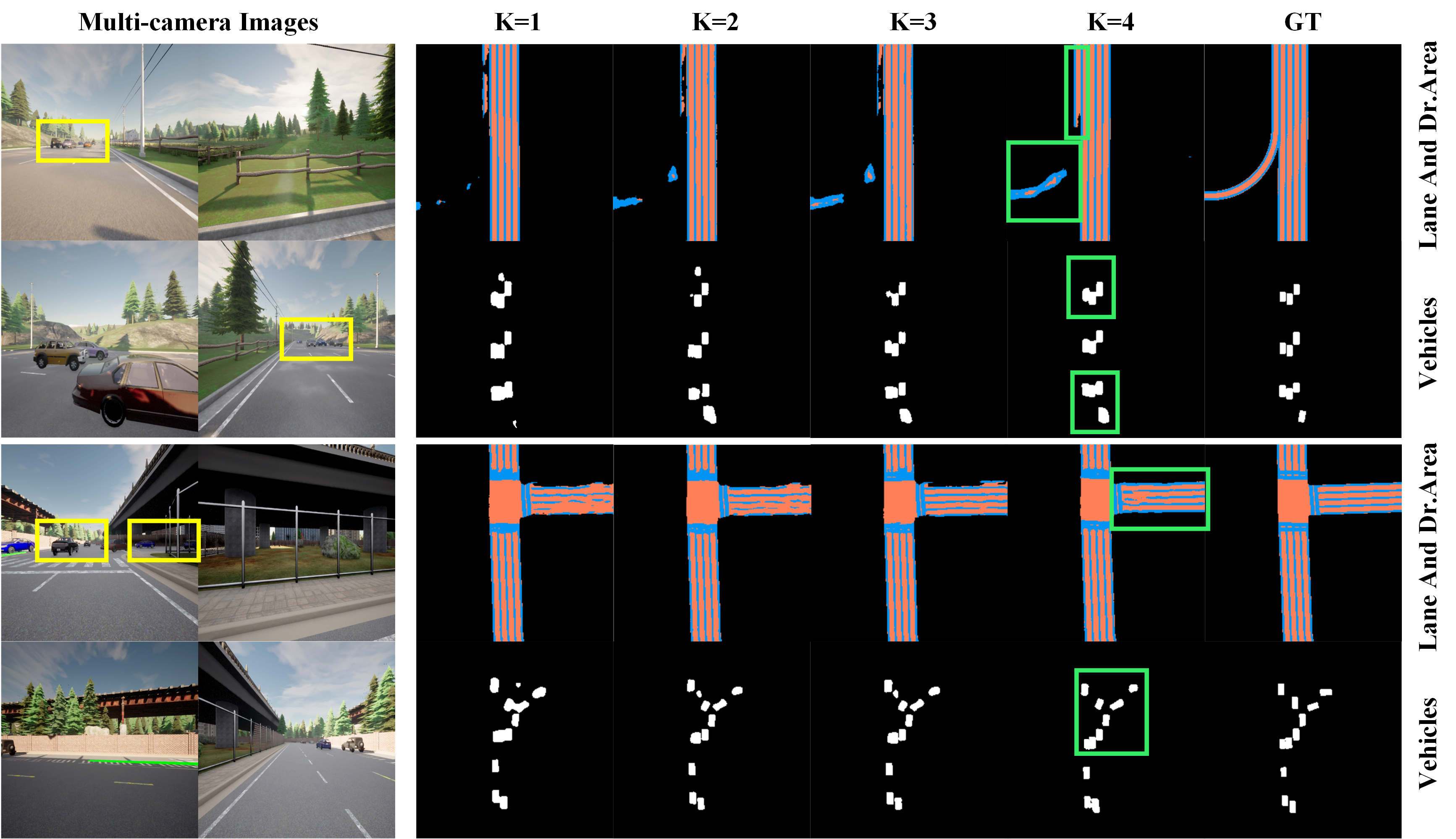}  
\end{center}
\caption{Visualization of the BEV graph generated after each normalizing flow step $K$ in CFLOW. We performed a qualitative analysis of $K$ in the OPV2V dataset, where the number of steps for each $K$ contains visualisations of three categories (vehicle, lane and drivable area). The yellow boxes represent the vehicle in the remote view and the obscured road. The green boxes represent the fine features that are restored after the flow transformation.}
\label{Fig8}
\end{figure*}

\subsection{Ablation Study}
In this section, we performed ablation experiments on various modules of TVB to analyze their effectiveness. We explored the differences between conditional variational inference and general variational inference, the impact of adding normalizing flow after conditional variational inference, and the influence of $K$-step flow transformations. We also analyzed the number of samples used to generate BEV maps. Finally, we investigated the importance of the BAF module for the overall framework.

\begin{table}[t]
\renewcommand\arraystretch{1.6}
\caption{Comparison of results of different methods on OPV2V datasets. All fusion methods are based on the CVT \cite{zhou2022cross} backbone.}
\begin{center}
\setlength{\tabcolsep}{3.5mm}{
\begin{tabular}{c|ccc}
\toprule
\multirow{2}{*}{Method} & \multicolumn{3}{c}{IoU}                               \\  
                        & \multicolumn{1}{c}{Vehicles} & \multicolumn{1}{c}{Drivable Area} & Lane \\ \hline
CVT \cite{zhou2022cross}& \multicolumn{1}{c}{50.1}     & \multicolumn{1}{c}{60.0}          & 44.1 \\ 
F-Cooper \cite{chen2019f}& \multicolumn{1}{c}{52.5}     & \multicolumn{1}{c}{60.4}          & 46.5 \\ 
AttFuse \cite{xu2022opv2v}& \multicolumn{1}{c}{51.9}     & \multicolumn{1}{c}{60.5}          & 46.2 \\ 
V2VNet \cite{wang2020v2vnet}& \multicolumn{1}{c}{53.5}     & \multicolumn{1}{c}{60.2}          & 47.5 \\ 
DiscoNet \cite{li2021learning}& \multicolumn{1}{c}{52.9}     & \multicolumn{1}{c}{60.7}          & 45.8 \\ \hline 
Ours                   & \multicolumn{1}{c}{\textbf{56.8}}     & \multicolumn{1}{c}{\textbf{61.3}}          & \textbf{48.1} \\ \bottomrule
\end{tabular}}
\end{center}
\label{tab2}
\end{table}
1) \textit{\textbf{Impact of ${{\mathcal{L}}_{KL}}$ loss and ${{\mathcal{L}}_{r}}$ reconstruction loss.}} We examined the performance of TVB w/ or w/o two loss functions on the nuScenes. In Table \ref{tab3}, the first row shows the results when both loss functions are absent, equivalent to an unconstrained convolutional segmentation network. The second row introduces the ${{\mathcal{L}}_{r}}$ loss, which acts as an intermediate supervisory term for the whole network, ensuring that the generated maps contain basic information such as vehicles and lanes. In the third row, both losses are applied in the network. As the prior network learns the posterior information of the posterior network, the ${{\mathcal{L}}_{KL}}$ loss makes the generated maps more targeted. 
\begin{table}[t]
\renewcommand\arraystretch{1.6}
\caption{Quantitative analysis of model performance by ${{\mathcal{L}}_{r}}$,${{\mathcal{L}}_{KL}}$ and normalizing flows on nuScenes val set.}
\begin{center}
\setlength{\tabcolsep}{3.2mm}{
\begin{tabular}{ccc|cc}
\toprule
${{\mathcal{L}}_{r}}$ & ${{\mathcal{L}}_{KL}}$ & FLOW & Setting 1 & Setting 2 \\ \hline
\xmark   &\xmark    &\xmark      & 38.5    & 37.4     \\ 
\Checkmark   &\xmark    &\xmark      & 40.3    & 38.9     \\ 
\Checkmark   &\Checkmark    &\xmark      & 41.8    & 40.1    \\ 
\Checkmark   &\Checkmark    &\Checkmark      & \textbf{43.2}    & \textbf{41.3} \\ \bottomrule
\end{tabular}}
\end{center}
\label{tab3}
\end{table}
\begin{table}[t]
\renewcommand\arraystretch{1.6}
\begin{center}
\caption{\textcolor{red}{EFFECT OF CFLOW STEPS $K$ ON TVB IN OPV2V.}}
\setlength{\tabcolsep}{3.2mm}{
\begin{tabular}{c|ccc|c|c}
\toprule
\multirow{2}{*}{$K$} & \multicolumn{3}{c|}{IoU}                                             & \multirow{2}{*}{\#Params} & \multirow{2}{*}{FPS} \\ \cline{2-4}
                   & \multicolumn{1}{c|}{Vehicles} & \multicolumn{1}{c|}{Dr. Area} & Lane &                         &                      \\ \hline
1                  & \multicolumn{1}{c|}{53.1}     & \multicolumn{1}{c|}{60.1}     & 46.2 & 6.2                     & 27                   \\
2                  & \multicolumn{1}{c|}{54.6}     & \multicolumn{1}{c|}{60.5}     & 46.9 & 6.4                     & 23                   \\
3                  & \multicolumn{1}{c|}{55.9}     & \multicolumn{1}{c|}{60.9}     & 47.5 & 6.9                     & 21                   \\
4                  & \multicolumn{1}{c|}{\textbf{56.8}}     & \multicolumn{1}{c|}{\textbf{61.3}}     & 47.7 & 7.8                     & 19                   \\
5                  & \multicolumn{1}{c|}{56.7}     & \multicolumn{1}{c|}{61.2}     & \textbf{48.1} & 9.4                     & 14                   \\ 
6                  & \multicolumn{1}{c|}{56.1}     & \multicolumn{1}{c|}{60.9}     & 47.8 & 12.0                    & 6                    \\  \bottomrule
\end{tabular}}
\end{center}
\label{tab4}
\end{table}
\begin{figure*}[t!]  
\begin{center}
   \includegraphics[width=1\linewidth]{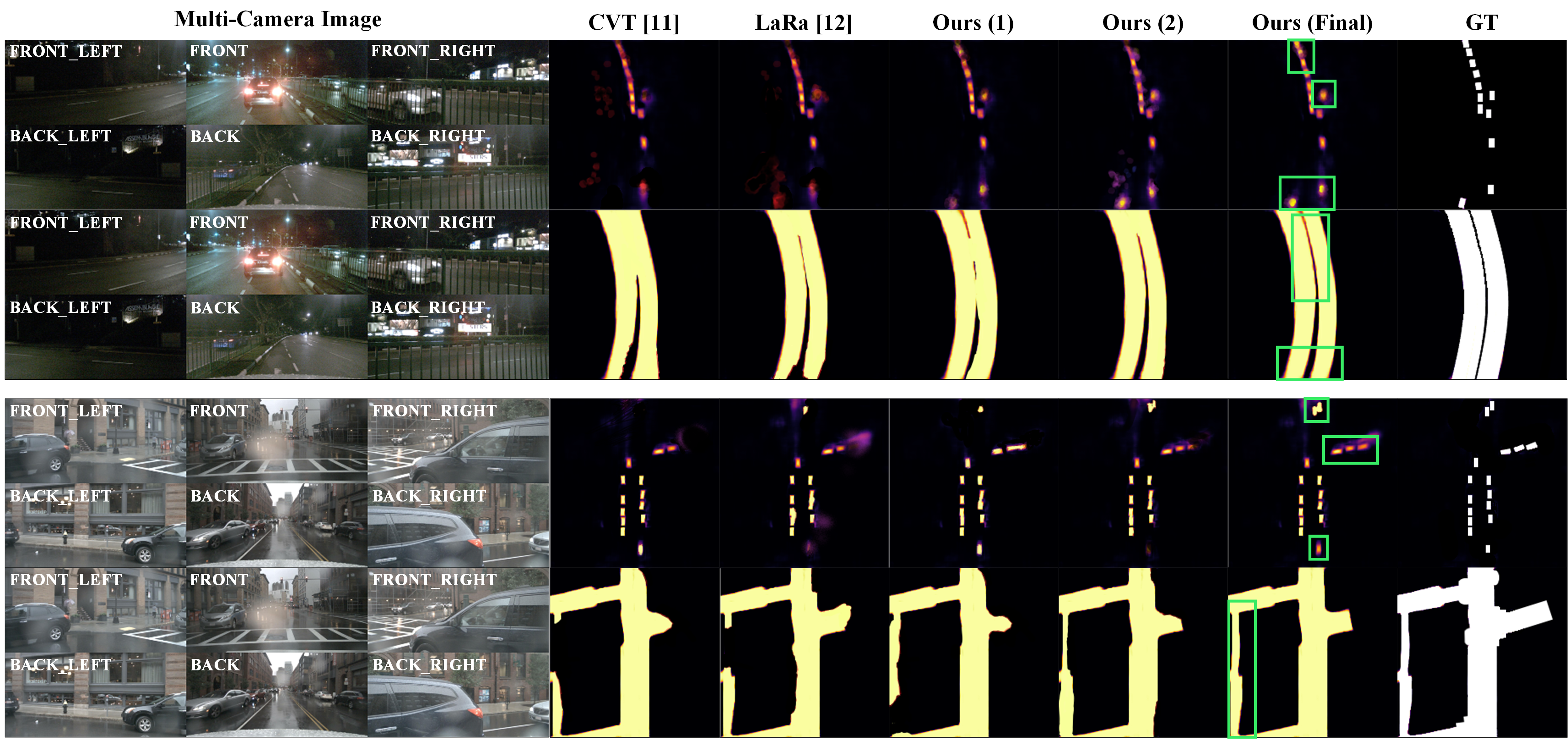}  
\end{center}
\caption{Visualization results of TVB in challenging scenarios. We present the visualization results of TVB in night and rainy scenarios, focusing on two categories: vehicles and drivable areas. We compare these results with those of CVT \cite{zhou2022cross} and LaRa \cite{bartoccioni2023lara}. “Ours (1)” and “Ours (2)” represent two of the multiple candidate maps generated by TVB, while “Ours (Final)” is the result obtained through fusion of multiple candidate maps.}
\label{Fig9}
\end{figure*}

2) \textit{\textbf{Impact of normalizing flow}}. To assess this, we examined the influence of normalizing flow on the model. As shown in Table \ref{tab3}, the last row indicates a significant improvement in TVB's IoU after incorporating normalizing flow, suggesting that normalizing flow plays a crucial role in generating BEV maps. While the CVAE model demonstrates strong representational capabilities in generating BEV maps, we aimed to further refine the information on lanes and vehicles. Normalizing flow enhances the expressiveness of the simple prior base distribution by introducing non-linearity, allowing more realistic BEV segmentation results through the sampling and reconstruction of the transformed distribution. This is particularly important in challenging scenarios.

3) \textit{\textbf{Impact of the number of CFLOW transformation steps ($K$)}}. Using a sequence of bijective transformations (flow chain) in the CFLOW module, the normalizing flow alters a simple base distribution and transforms it into a complex and realistic target distribution. Throughout the process, the latent distribution is transformed into the target distribution by executing $K$ transformation steps. We present visualizations of each step of $K$ transformations for the categories of vehicles, lane and driveable area, as shown in Fig.~\ref{Fig8}. As the number of $K$ steps increases, the IoU for both categories gradually improves, resulting in more accurately generated BEV segmentation, as shown in Table IV. However, considering the potential trade-off between model performance and computational cost, we observe diminishing returns in performance improvement as $K$ increases, which may lead to overfitting. In Table IV, we show that the number of network parameters (Params) and the Frames Per Second (FPS) of TVB increase with the number of $K$ steps. Therefore, we opt for a balance between model speed and performance by choosing $K=4$.
\begin{table}[t]
\renewcommand\arraystretch{1.6}
\begin{center}
\caption{The performance comparison of BAF and MC modules on the OPV2V val set by TVB. Here, MC stands for Monte Carlo sampling method.}
\setlength{\tabcolsep}{3.2mm}{
\begin{tabular}{c|ccc}
\toprule
\multirow{2}{*}{Module} & \multicolumn{3}{c}{IoU}                                                   \\ \cline{2-4} 
                        & \multicolumn{1}{c|}{Vehicles} & \multicolumn{1}{c|}{Driveable Area} & Lane \\ \hline
MC                      & \multicolumn{1}{c|}{53.6}     & \multicolumn{1}{c|}{60.1}           & 46.4 \\
BAF                     & \multicolumn{1}{c|}{\textbf{56.8}}     & \multicolumn{1}{c|}{\textbf{61.3}}           & \textbf{48.1} \\\bottomrule
\end{tabular}}
\end{center}
\label{tab6}
\end{table}
\begin{table}[t]
\renewcommand\arraystretch{1.6}
\begin{center}
\caption{The results of TVB on two datasets under different sampling numbers. The results represent the average values across all categories in each dataset.}
\setlength{\tabcolsep}{3.2mm}{
\begin{tabular}{c|ccccc}
\toprule
\# Samples & 1 & 10 & 30   & 50 & 100 \\ \hline
nuScenes      & 40.2  & 40.9   & \textbf{41.3} & 41.2   & 41.2    \\ 
OPV2V      & 54.0  & 54.5   & \textbf{55.4} & 55.3   & 55.0    \\ \bottomrule
\end{tabular}}
\end{center}
\label{tab7}
\end{table}

4) \textit{\textbf{Impact of the BAF module}}. Table V shows the impact of the BAF module on the model performance. After generating multiple BEV maps with CFLOW, we did not perform simple averaging on them, as there may be some potentially correct samples within them. We want to maximize the contribution of relatively correct samples to improve the final BEV segmentation results, and conversely, to weaken the contribution of negative samples. In the BAF module, we consider global features for all samples and assign weights to them. We then focus on the local features of the samples to further improve the detailed features. Finally, BAF adaptively fuses multiple BEV maps to achieve accurate BEV segmentation.

5) \textbf{\textit{Impact of the number of generated BEV maps.}} We sample multiple latent variables by transforming the latent prior distribution through a flow and feed them into the decoder to reconstruct multiple candidate BEV maps. We conduct TVB by collecting different numbers of samples on two datasets, and the results in Table VI represent the mean IoU across all classes in each dataset. The performance of the model decreases when the number of samples exceeds 50, possibly indicating the onset of overfitting. Furthermore, as the sample size increases, the lightweight attention mechanism may struggle to effectively capture differences between various samples, leading to a decrease in performance. Therefore, we set the sample size at 30 to ensure the effectiveness of TVB.

\subsection{Discussion on Efficiency and Limitations}
Although TVB achieves competitive BEV segmentation performance on both nuScenes and OPV2V, its efficiency and fine-structure modeling remain limited. During inference, TVB draws 30 Monte Carlo samples to generate candidate BEV maps, which are then adaptively fused by the BAF module. This sampling-and-fusion strategy improves robustness but also introduces extra computational overhead, leading to 19 FPS on nuScenes, as reported in Table \ref{tab1}. The cost further increases with the number of normalizing flow steps $K$. While a larger $K$ enhances the expressiveness of the latent distribution, it reduces inference speed substantially; for example, the FPS drops to 6 when $K$=6. We therefore adopt $K$=4 as a practical trade-off between accuracy and efficiency.

The improvement is also relatively limited for categories requiring high geometric precision, such as lane segmentation. On OPV2V, TVB achieves 47.7 IoU for the Lane category, where the gain is less pronounced than that for larger region-level categories such as Vehicles and Drivable Area. This suggests that the current generative framework is less effective in modeling thin and elongated structures. Future work will explore geometry-aware constraints, boundary supervision, and more efficient sampling strategies to improve real-time deployment capability and fine-structure segmentation accuracy.

\section{Conclusion}
In this paper, we transformed the BEV segmentation problem into a generative problem within the variational flow framework and proposed a variational BEV segmentation network based on the transformer. By introducing normalizing flow into the conditional variational framework, the latent prior distribution became more complex, and multiple candidate BEV maps were obtained through repeated sampling, resulting in more realistic representations. Our TVB utilized the proposed BAF module to fuse multiple BEV maps and obtain the final BEV segmentation. Experimental results on two benchmark datasets indicated that the performance of our TVB surpassed that of the previous BEV segmentation methods. Nevertheless, the proposed sampling-based generative framework introduces additional inference overhead, especially when multiple Monte Carlo samples and larger normalizing-flow steps are used. Moreover, its advantage is relatively limited for thin structures such as lanes. In future work, we will investigate more efficient sampling strategies and geometry-aware constraints to improve real-time deployment capability and fine-structure segmentation accuracy.

\bibliography{name}
\bibliographystyle{IEEEtran}

\vfill

\end{document}